\documentclass[a4paper]{article}

\usepackage[english]{babel}
\usepackage[utf8x]{inputenc}
\usepackage[T1]{fontenc}
\usepackage{ dsfont }
\usepackage{placeins}
\usepackage{multirow}
\usepackage{tabularx} 
\usepackage{lscape}
\usepackage{algpseudocode}
\usepackage{color,soul}
\usepackage{bigfoot}
\usepackage{csvsimple}
\usepackage{longtable}
\usepackage{arydshln,amssymb,color}
\usepackage{multicol}
\usepackage{bm}
\usepackage{tikz}
\usetikzlibrary{arrows.meta}
\usepackage{amsmath}
\usepackage{graphicx}
\usepackage[colorinlistoftodos]{todonotes}
\usepackage{booktabs} 
\usepackage{xtab,booktabs}
\usepackage{footnote}
\makesavenoteenv{tabular}
\makesavenoteenv{table}
\usepackage{ dsfont }
\usepackage{ stmaryrd }
\usepackage{flushend}
\usepackage{natbib}
\usepackage{xspace}
\usepackage[paper=a4paper,margin=0.7in]{geometry} 

\usepackage{subcaption}

\usepackage[linesnumbered,algoruled,boxed,lined,noend]{algorithm2e}
\SetKwInOut{KwParam}{Parameters}
\usepackage{booktabs} 

\definecolor{customdarkblue}{HTML}{444683}

\newcommand\evogym{Evogym\xspace}
\newcommand\robogrammar{Robogrammar\xspace}

\newcommand\veenstra{gym-rem 2D\xspace}
\newcommand\jorgen{gym-rem\xspace}

\newcommand\nenvironments{four\xspace}

\usepackage{xcolor}
\definecolor{myblue}{RGB}{0, 107, 179}
\definecolor{mygreen}{RGB}{0, 128, 0}
\usepackage[colorlinks=true,linkcolor=myblue,citecolor=mygreen,urlcolor=myblue,hyperfootnotes=false]{hyperref}
\hypersetup{
	linkbordercolor=myblue,
	citebordercolor=mygreen,
	urlbordercolor=myblue,
	pdfborderstyle={/S/U/W 1}
}

\title{
	An Empirical Study on the Computation Budget of Co-Optimization of Robot Design and Control in Simulation
}
\author{Etor Arza$^{1}$ \and Frank Veenstra$^{2}$ \and Tønnes F. Nygaard$^3$ \and Kyrre Glette$^{45}$}
\date{
\small
    \texttt{etorarza@gmail.com}\\%
    $^1$NTNU Institutt for teknisk kybernetikk, Norway\\%
    \texttt{frank.veenstra@inn.no}\\%
    $^2$Department of Game Development - Inland Norway University\\%
    \texttt{tonnesfn@uio.no}\\%
  	$^3$Department of Technology Systems, University of Oslo, Norway \\%
    \texttt{kyrrehg@ifi.uio.no}\\%
    $^4$RITMO Centre for Interdisciplinary Studies in Rhythm, Time and Motion, University of Oslo, Norway \\%
    $^5$Department of Informatics, University of Oslo, Norway \\%
}
\begin{document}
	\maketitle
	
	\begin{abstract}
		The design (shape) of a robot is usually decided before the control is implemented.
		This might limit how well the design is adapted to a task, as the suitability of the design is given by how well the robot performs in the task, which requires both a design and a controller.
		The co-optimization or simultaneous optimization of the design and control of robots addresses this limitation by producing a design and control that are both adapted to the task.
		This paper investigates some of the challenges inherent in the co-optimization of design and control in simulation.
		The results show that reducing how well the controllers are trained during the co-optimization process significantly improves the robot's performance when considering a second phase in which the controller for the best design is retrained with additional resources.
		In addition, the results demonstrate that the computation budget allocated to training the controller for each design influences design complexity, with simpler designs associated with lower training budgets.
		This paper experimentally studies key questions discussed in other works in the literature on the co-optimization of design and control of robots in simulation in four different co-optimization problems.


	\end{abstract}

	
	\section{Introduction}
	\label{section:intro}
	
The behavior of a robot is tied to its design.
In fact, the design defines (sometimes by itself) how the robot is going to move.
An early example of this is McGeer's \textit{Passive Dynamic Walking} robot~\citep{mcgeerPassiveDynamicWalking1990}.
This robot has a ``human-like'' walking behavior when on a slope, but requires no energy nor control. 
Of course, robots with no control have a limited ability to interact with complex environments.
Hence, both the design and control play a key role in the behavior of robots.

Usually, the design of the robots is carried out by hand before the controller is even implemented.
This is done in an attempt to make the design suitable for the task.
Therefore, the control is considered after the design has already been chosen.
However, this might not always be the optimal approach, as it is difficult to judge how good the design of a robot is without a suitable controller that can properly take advantage of the design~\citep{rosendoTradeoffMorphologyControl2017}.
In addition, hand designing is a slow and labor-intensive process~\citep{liaoDataefficientLearningMorphology2019}, and alternatively, it is possible to jointly optimize the design and control of robots.
While some approaches have been proposed to co-optimize design and control directly in the real world \citep{nygaardRealworldEmbodiedAI2021,rosendoTradeoffMorphologyControl2017}, the co-optimization is usually carried out in simulation due to the high cost of building multiple robots.
In this paper, we focus on the co-optimization of design and control of robots in simulation.

\begin{figure}
	\centering
	\includegraphics[width=0.81\linewidth]{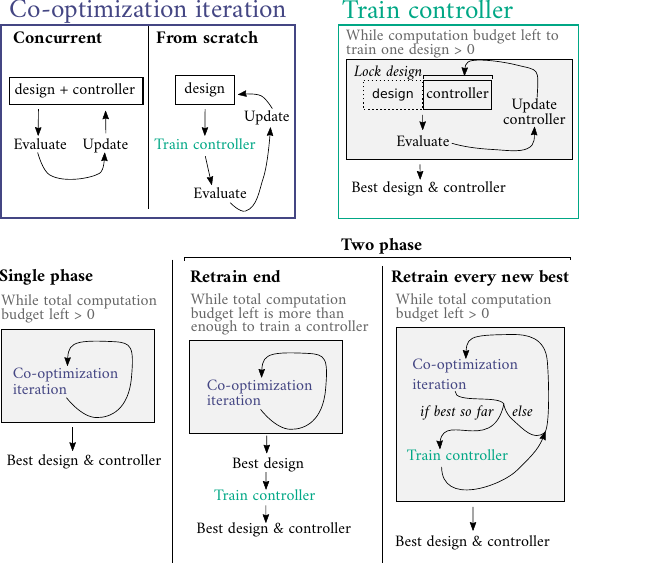}
	\caption{
     Different co-optimization approaches. 
     Each iteration in the co-optimization process can update the design and control together (\textbf{Concurrent}), or train a controller from scratch for each design update (\textbf{From scratch}).
     On a higher level, the most common co-optimization approach is \textbf{Single phase}, in which both the controller and design are optimized together. 
     It is also possible to retrain the controller for the best designs with additional resources.
     In the case of \textbf{Two phase: retrain end}, the additional training procedure is carried out directly after the co-optimization approach terminates; while \textbf{Two phase: retrain every new best} retrains each new best design found during the co-optimization process.
	}
	\label{fig:nestedvsconcurrentdefinition}
\end{figure}

The co-optimization of design and control of a robot is an iterative optimization process where both the design and the control of the robot are improved together.  
In general, there are two method types~\citep{hartArtificialEvolutionRobot2022} depending on how the control of the designs is improved in each iteration: it can be done \textit{concurrently} (also known as \textit{inheritance of control} by some authors~\citep{luo2023lamarcksrevengeinheritancelearned}) or \textit{from scratch} (see \color{customdarkblue}Co-optimization iteration \color{black} in Figure~\ref{fig:nestedvsconcurrentdefinition}).
The \textit{concurrent} co-optimization involves simultaneously co-optimizing the design and control, where the controller for each new design is initialized with the controller of the previous design.
This is the oldest~\citep{sims1994evolving,lipsonAutomaticDesignManufacture2000} and most used approach due to its computing efficiency~\citep{10.1162/isal_a_038}.

However, \textit{concurrent} approaches have a few limitations.
Firstly, the design tends to converge before the controller, which can lead to a non-optimal design~\citep{lipsonDifficultyCooptimizingMorphology2016}.
When a design change is introduced, it tends to worsen the objective value, as the controller has not learned to deal with this change yet~\citep{cheneyScalableCooptimizationMorphology2018,nygaardOvercomingInitialConvergence2017,faina2013edhmor}. 
A potential solution is to give some time for the controller to improve before concluding how good the design is~\citep{cheneyScalableCooptimizationMorphology2018}.
Additionally, \textit{concurrent} requires that controllers are compatible with different designs, which can limit the design space~\citep{legoffMorphoevolutionLearningUsing2021,hartArtificialEvolutionRobot2022}. 
For example, if two designs have a different number of sensors and actuators, a controller will not necessarily be compatible with both.
And even when the controllers are compatible between designs, it is not always the case that the performance is good when transferring the control from one design to another, specially when they differ significantly.

The alternative to the concurrent approach is to learn the controller from scratch for each new design that is evaluated (denoted as \textit{from scratch}~\citep{hartArtificialEvolutionRobot2022}, as a shortened version of \textit{train the controller from scratch for each design}).
\textit{From scratch} procedures instead train a controller from scratch for each design update (see \color{customdarkblue}Co-optimization iteration \color{black} in Figure~\ref{fig:nestedvsconcurrentdefinition}).
This iteratively improves the design, where the objective value of the design is computed given a controller specifically trained for that design: Evaluating each design requires training a new controller from scratch, and then evaluating the design with the controller.
The computation budget of the co-optimization process is therefore given mostly by the training procedure for the controller, which is executed for each controller trained and usually involves significantly more computation than evaluating a controller and a design.

\textit{From scratch} methods overcome the challenges of \textit{concurrent} methods mentioned above.
They are also easier to apply, as they can use existing control training procedures directly with little modifications to set-up a co-optimization of design and control problem.
The main disadvantage of \textit{from scratch} procedures is the increase in required computation budget: the co-optimization problem essentially becomes a nested optimization problem, where each evaluation of a design first requires optimizing the controller for the design.
There have been some attempts to overcome the inefficiency of \textit{from scratch} approaches, such as initializing the controller to the best found controller so far that is compatible with the current design~\citep{legoffMorphoevolutionLearningUsing2021} or using distributed computing~\citep{guptaEmbodiedIntelligenceLearning2021a}.
However, the computational cost of the \textit{from scratch} method remains high.

With the \textit{from scratch} method, and given a fixed amount of computation budget, there is a trade-off between how well the design is trained versus how well the controller is trained.
For each design that is evaluated, the controller needs to be learned first.
The more resources we use to learn the controller for a particular design, the better that design will perform, but at a cost of a higher computational cost.
This higher cost implies that fewer designs can be evaluated within the same total computation budget.
This trade-off is associated with \textit{from scratch} approaches, and is not relevant for \textit{concurrent} methods.
Hence, this paper considers only \textit{from scratch} methods.

\begin{figure}
	\centering
	\includegraphics[width=\linewidth]{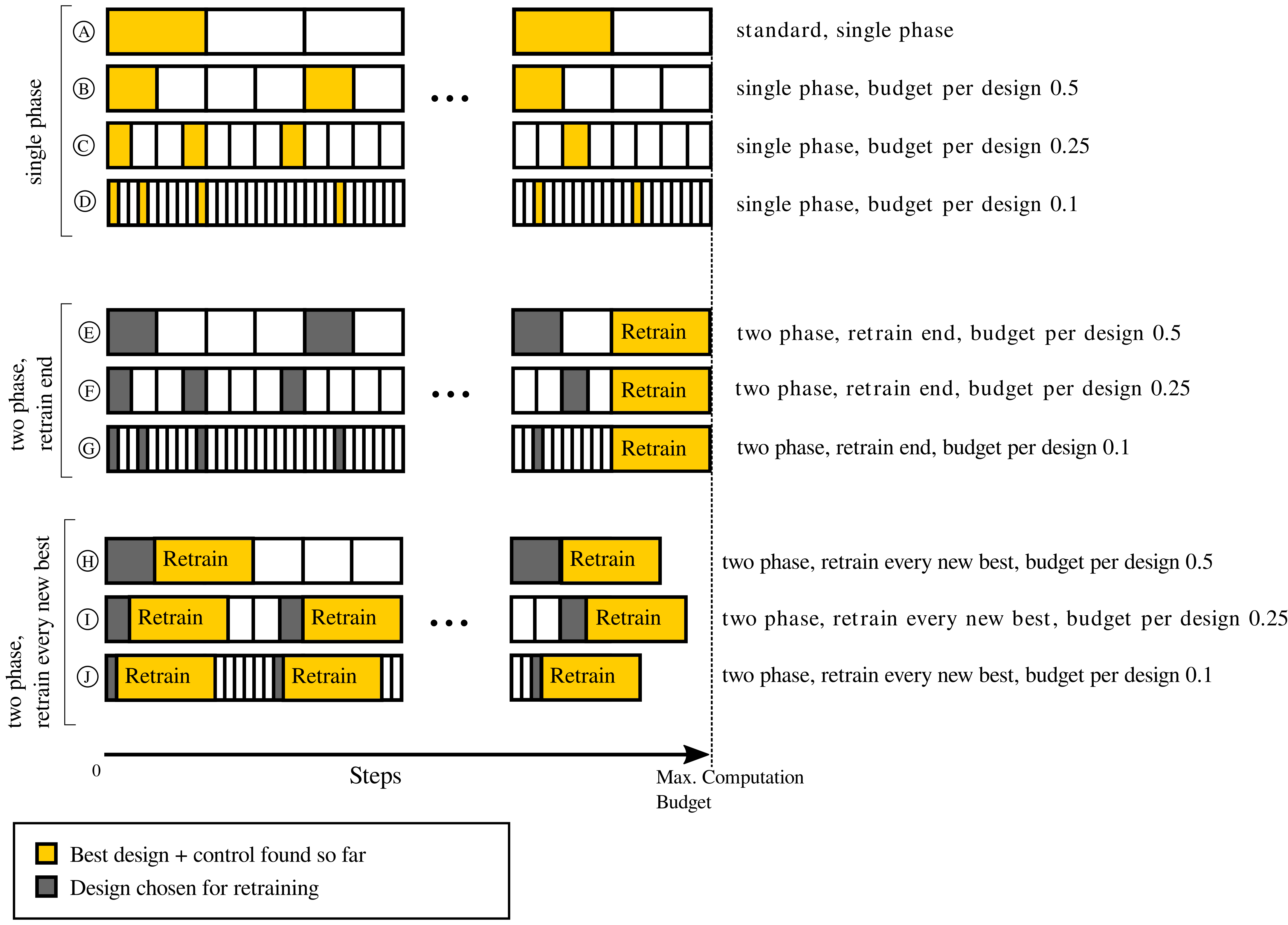}
	\caption{How each of the methods spend the same computation budget. Each box represents training the controller for a design. The length of each box is proportional to the number of steps used to train that controller.\textit{Two phase} approaches train controllers for different designs and choose the best designs (in gray) to be retrained with additional resources. \textit{Retrain every new best} retrains a design every time it finds it outperforms the previous best with a smaller computation budget. In contrast, \textit{retrain end} only retrains the best found design at the end of the optimization process.\\ 
	}
	\label{fig:computationalbudgetbarcharts}
\end{figure}

\pagebreak[3]
\subsection*{Related work}

\nopagebreak[4]
\emph{Retraining the best design}
\nopagebreak[4]

Given the computational complexity of the co-optimization process, different schemes have been proposed to allocate a given fixed computation budget.
The simplest allocation is to use the same computation budget of every design, which we denote as \textit{single-phase} in this paper (see bottom left in Figure~\ref{fig:nestedvsconcurrentdefinition}).
In addition, it is also possible to consider a two phase approach, where the controller of the best design is retrained with additional resources after the co-optimization procedure ends (see \textit{two phase, retrain end} in Figure~\ref{fig:nestedvsconcurrentdefinition}).

\citet{rosendoTradeoffMorphologyControl2017} compare a \textit{single phase} and a \textit{retrain end} approach.
They conclude that interrupting the co-optimization and continuing only with the optimization of control (\textit{retrain end}) is detrimental to the performance of the co-optimization, and that it is better to stick to a \textit{single phase} approach.
\citet{lipsonDifficultyCooptimizingMorphology2016} also conducted a similar experiment and reached a similar conclusion.
In contrast, \citet{nygaardOvercomingInitialConvergence2017}'s work shows that \textit{retrain end} outperforms \textit{single phase}.

\pagebreak[3]
\emph{Design complexity}
\nopagebreak[4]

Design complexity has different definitions in different works, but usually it is a measurement of the size of the representation of the design.
\citet{cheneyScalableCooptimizationMorphology2018} experimented on the co-optimization of voxel-based modular robots.
They observed both better performance and a higher design complexity in the best-performing designs when additional training budget is allocated to new designs.
\citet{lipsonDifficultyCooptimizingMorphology2016} also found that more challenging tasks produced more complex designs.
\citet{auerbachEnvironmentalInfluenceEvolution2014} observed that the complexity of the best-performing design increases with the difficulty of the environments.

Unlike the three previous works, \citet{10.1007/978-3-030-16692-2_41} and \citet{bongard2010utility} proposed co-optimizing design and control with a constant design complexity (design complexity does not change during co-optimization).
They execute independent co-optimization runs with different design space complexities and observed that, more complex design spaces outperform simpler designs spaces on more difficult tasks.
However, while \citet{10.1007/978-3-030-16692-2_41} observed an improvement in smaller design spaces for simpler tasks, complex design spaces performed as well as simpler ones on easier tasks in \citet{bongard2010utility}'s work.

\subsection*{Contributions}

In this paper, we study some of the challenges inherent in \textit{from scratch} co-optimization of design and control problems, and we try to answer two key questions associated with the related work mentioned before.
\begin{enumerate}
    \item Does a two-phase approach with an initial co-optimization phase followed by a controller-only optimization phase outperform a single-phase co-optimization approach?
    \item How does the design complexity change with respect to the amount of computational resources?
\end{enumerate}
To answer these two questions, we conduct three experiments in \nenvironments co-optimization environments available in the literature.
In the following, we describe the contributions of this paper in detail.

\pagebreak[3]
\underline{Contribution 1:}

\nopagebreak[4]
Previous work in the literature disagrees on whether an initial \textit{co-optimization} phase followed by a \textit{controller optimization only} phase is a better strategy than a single \textit{co-optimization} strategy.
Some works suggest that a single-phase strategy is better~\citep{rosendoTradeoffMorphologyControl2017,lipsonDifficultyCooptimizingMorphology2016}, while the work of \citet{nygaardOvercomingInitialConvergence2017} suggests that the two-phase approach can be better when the controller is reinitialized before the second phase.

An experimentation on \nenvironments different co-optimization environments validates the statement of \citet{nygaardOvercomingInitialConvergence2017}: we show that a two phase approach with an initial \textit{from scratch} co-optimization followed by a second phase in which the best design is retrained with additional budget outperforms the baseline single-phase co-optimization procedure.

\pagebreak[3]
\underline{Contribution 2:} 

\nopagebreak[4]
Two-phased approaches as in the works of \citet{rosendoTradeoffMorphologyControl2017}, \citet{lipsonDifficultyCooptimizingMorphology2016}, and \citet{nygaardOvercomingInitialConvergence2017} consider a first phase of co-optimization procedure followed by a second phase in which only the controller of the best-found design is optimized (\textcircled{\scriptsize E}, \textcircled{\scriptsize F}, \textcircled{\scriptsize G} in Figure~\ref{fig:computationalbudgetbarcharts}).
However, it is also possible to retrain the controller each time a new best design is found (\textcircled{\scriptsize H}, \textcircled{\scriptsize I}, \textcircled{\scriptsize J} in Figure~\ref{fig:computationalbudgetbarcharts}).
This ensures that the best design is chosen given a controller that is as good as possible for that design, but requires more computation budget.
We compare these two approaches and show that \textit{Two phase: retrain end} is a better approach when the computation budget is very low, while \textit{Two phase: retrain every new best} is preferable for larger computation budgets.

\pagebreak[3]
\underline{Contribution 3:} 

\nopagebreak[4]

\citet{mirasEvolvingcontrollersLearningcontrollersMorphologically2020} and later \citet{legoffMorphoevolutionLearningUsing2021} \citet{luoEffectsLearningMorphologically2022} observed that co-optimization improves the capability of the designs to learn a controller. 
In other words, designs observed at the end of the co-optimization process are better at learning the controller.
\citet{10.1007/978-3-030-16692-2_41} tested different control spaces to train gait controllers of a four-legged robot and found that a simpler (smaller) control space works better when the training budget is low.

We look at the complexity of the designs with respect to the training budget for each controller, and we show that, in most cases, a low training budget for each design during the co-optimization procedure tends to produce simpler designs (that are consequently easier to train).

\section{Experimentation}

In this section, we present three experiments in which we study some of the challenges inherent in the co-optimization of design and control and address the three contributions mentioned above.
We begin by introducing certain concepts that are required to understand and interpret the experimentation and the environments in which the experiments are carried out.

\setcounter{subsection}{-1}
\subsection{Experimental Framework}

\subsubsection*{General concepts}
\label{sec:general_definitions}

In the following, we define a few concepts that are essential for understanding the experimentation and interpreting the results.

$\bullet$ \emph{computation budget}.
The computation budget in this paper is measured in steps, which is the standard terminology considered in reinforcement learning~\citep{brockmanOpenAIGym2016}. 
Given an agent in a reinforcement learning environment, a step involves computing the next action of the policy and simulating this action in the environment, which are usually the two costliest parts of the co-optimization process.
A step in simulation usually equals anywhere between $0.1$ to $0.0001$ seconds of real-time in the environment that is simulated, depending on the fidelity required for the environment.
Changing the step size can change the behavior of the environment~\citep{echevarrieta2024speeding} and therefore, in this paper, we keep the original step size of the authors.
In contrast to time~\citep{arzaFairComparisonOptimization2024}, steps are transferable across different hardware, which makes the results more reproducible.
In any case, the most important thing to remember is that the number of computed steps is proportional to the time, and hence we use it as a measure of time in the rest of the paper.

$\bullet$ \emph{episode}.
An episode consists of initializing a design and a controller, and evaluating them in the environment for a fixed amount of steps.

%

$\bullet$ \emph{budget per design $\in (0, 1]$}.
Given a fixed computation budget, the number of designs that can be evaluated in that time is proportional to the computation budget considered to train each of the designs.
Hence, it is possible to stop training the designs earlier, and consequently, be able to evaluate additional designs with the same total computation budget.
We can achieve this by considering a lower stopping criterion in terms of steps for the training procedure of the controllers.
For example, if we reduce the budget to train the controller by 75\%, we are able to evaluate four times as many designs.
To simplify notation, we denote this as a \textit{budget per design} of $0.25$ (see \textcircled{\scriptsize C} in Figure~\ref{fig:computationalbudgetbarcharts}).

$\bullet$ \emph{two-phase, retrain end}.
The (\textit{single-phase}) co-optimization of design and control is an optimization algorithm that iteratively improves the design and control of a robot. 
However, it is possible to add a second phase in which the design is locked and only the control is optimized~\citep{rosendoTradeoffMorphologyControl2017,lipsonDifficultyCooptimizingMorphology2016,nygaardOvercomingInitialConvergence2017} (see \textcircled{\scriptsize E}, \textcircled{\scriptsize F}, \textcircled{\scriptsize G} in Figure~\ref{fig:computationalbudgetbarcharts}).
The second phase is executed after the co-optimization ends, and it further optimizes the controller of the best-found design.

%

$\bullet$ \emph{two-phase, retrain every new best}.
In contrast to the \textit{retrain end} approach, it is also possible to execute the second phase (which locks the design and only optimizes the control) every time a new best design is found. 
This implies reinitializing the controller and training it again with more resources, for every new best design found during the co-optimization process.
This approach uses more resources, as the retraining procedure is executed multiple times, but it ensures that the best-found design will be selected given a controller that is properly trained for that design (see \textcircled{\scriptsize H}, \textcircled{\scriptsize I}, \textcircled{\scriptsize J} in Figure~\ref{fig:computationalbudgetbarcharts}).

%

\subsubsection*{Four co-optimization environments}

\begin{figure}[htbp]
	\centering
	\begin{subfigure}[b]{0.35\textwidth}
		\includegraphics[width=\textwidth]{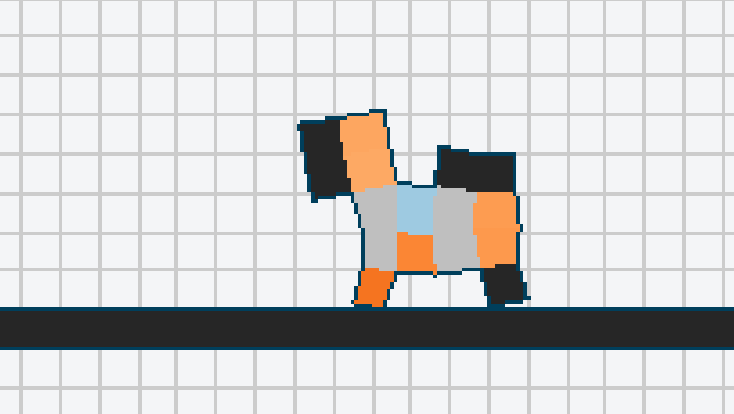}
		\caption{Evogym \citep{bhatia2021evolution}}
		\label{fig:evogym_screenshot}
	\end{subfigure}
	\begin{subfigure}[b]{0.35\textwidth}
		\includegraphics[width=\textwidth]{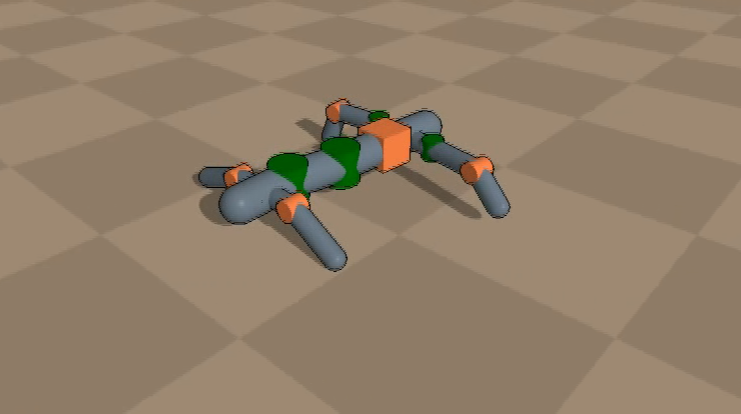}
		\caption{\robogrammar \citep{10.1145/3414685.3417831}}
		\label{fig:robogrammar_screenshot}
	\end{subfigure}
	
	\begin{subfigure}[b]{0.35\textwidth}
		\includegraphics[width=\textwidth]{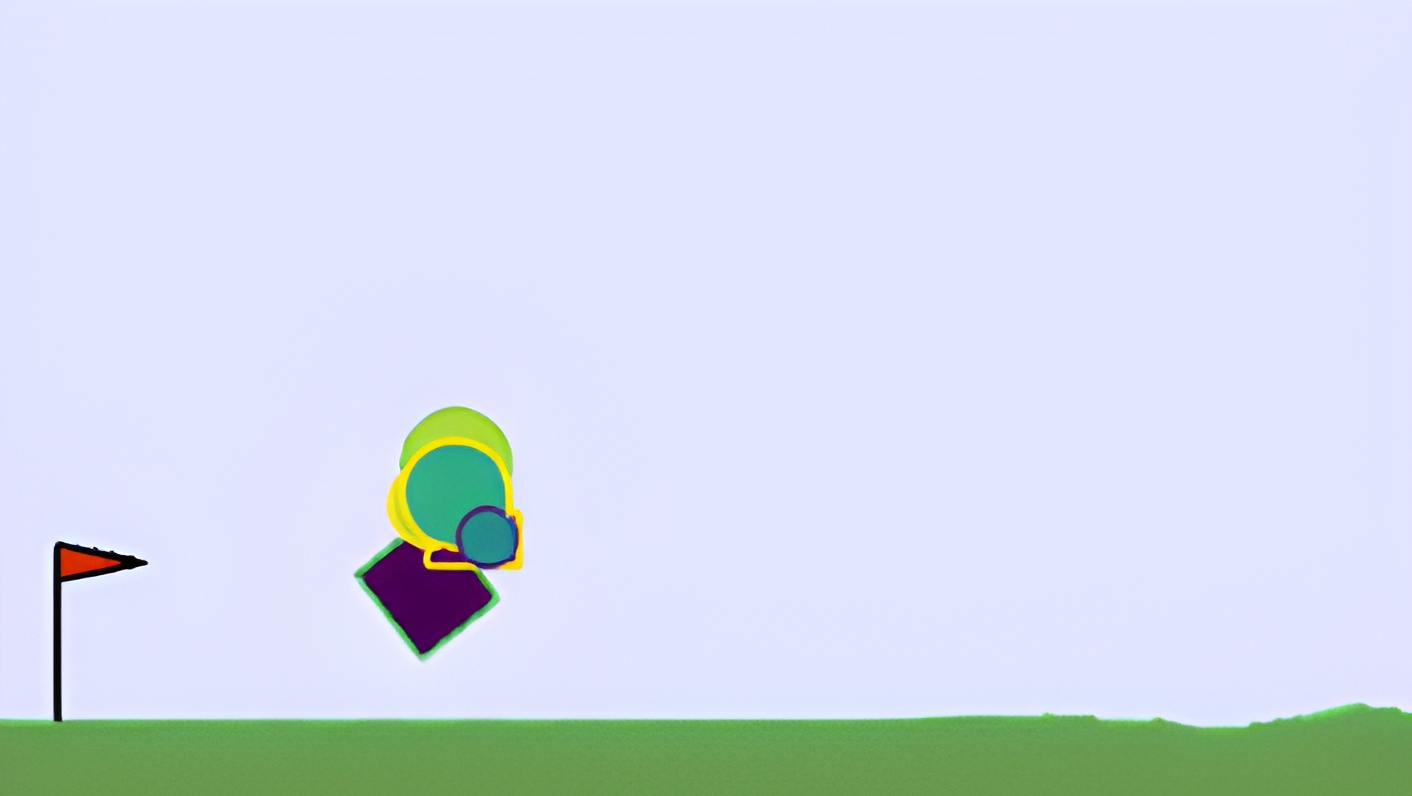}
		\caption{\veenstra \\ \ \ \citep{veenstra2020different}}
		\label{fig:gyrem2d_screenshot}
	\end{subfigure}
	\begin{subfigure}[b]{0.35\textwidth}
		\includegraphics[width=\textwidth]{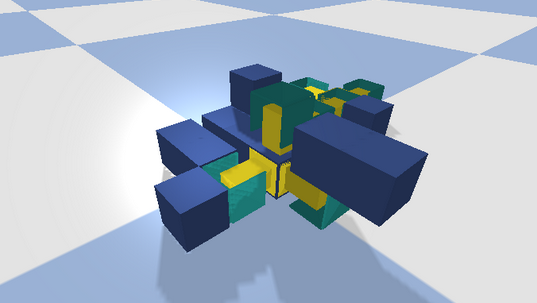}
		\caption{\jorgen \\ \ \ \citep{nordmoenMAPElitesEnablesPowerful2021}}
		\label{fig:jorgen_screenshot}
	\end{subfigure}

	\caption{The \nenvironments co-design environments considered in this paper.}
	\label{fig:screenchots_of_frameworks}
\end{figure}

In this paper, we experiment on \nenvironments environments from the literature in which the design and controller of robots are co-optimized.
In the following, we briefly introduce each of these environments.
A screenshot of each environment is shown in Figure~\ref{fig:screenchots_of_frameworks}.

\textbf{\evogym} by \citet{bhatia2021evolution} is an environment in which the design and controller of $5 \times 5$, 2D soft robots are learned for different tasks (Figure~\ref{fig:evogym_screenshot}).
The design is learned with genetic algorithms and the controller is learned with proximal policy optimization~\citep{schulman2017proximal}.
The environment includes many types of tasks like climbing, object manipulation, and movement.
In this paper, we consider the simplest task, which is to move to the right as far as possible within the given episode length.

\textbf{\robogrammar} is an environment by \citet{10.1145/3414685.3417831} that proposes learning the design and control of 3D legged robots (Figure~\ref{fig:robogrammar_screenshot}).
The process is carried out in two steps.
First, the grammar space is reduced to a meaningful subset.
Then, the design and control of the robots are co-optimized.
The objective function is the distance that the robots are able to move from the beginning.
In this paper, we focus on this second step.

\textbf{\veenstra} by \citet{veenstra2020different} is a comparison of different encoding methods for the \textit{concurrent} optimization of design and control (Figure~\ref{fig:gyrem2d_screenshot}).
The design and control of a 2D robot based on different interconnected shapes need to move to the right as far as possible.
The robot is controlled by the connections between the shapes, which contract and expand given some constant control parameters.
Considering that the \veenstra approach is concurrent, we proposed a \textit{from scratch} approach that learns a completely new controller via CMA-ES~\citep{augerTutorialCMAESEvolution2012} for each design evaluated.

\textbf{\jorgen} is an environment in which modular 3D robots are built out of cubes \citep{nordmoenMAPElitesEnablesPowerful2021}. 
The cubes are connected to each other, and the robot moves by rotating the cubes.
The design and control of the robots are learned with a \textit{concurrent} method.
In this paper, we considered this environment with a \textit{from scratch} approach, where we optimize the controller for each design from scratch with CMA-ES~\citep{augerTutorialCMAESEvolution2012}.

Each environment has a different optimization algorithm for the design and the control.
Table~\ref{tab:learning_algorithms} summarizes the optimization algorithms considered in these \textit{from scratch} approaches.

\renewcommand{\arraystretch}{1.2} 
\begin{table}
\centering
\begin{tabular}{ |p{2.5cm}|p{5.1cm}|p{4.5cm}| }
	\multicolumn{1}{c}{\ } & \multicolumn{1}{c}{Design} & \multicolumn{1}{c}{Control}\\ \cline{1-3}
	\evogym & Genetic Algorithm  \citep{10.5555/534133} & Proximal Policy Optimization~\citep{schulman2017proximal} \\ \cline{1-3}
	\robogrammar & Monte Carlo Tree Search~\citep{kocsis2006bandit} & Model Predictive Path Integral Control~\citep{lowrey2018plan} \\ \cline{1-3}
	\veenstra &  $\mu, \lambda$ Evolution Strategy~\citep{eiben2015introduction}  & CMA-ES~\citep{augerTutorialCMAESEvolution2012} \\ \cline{1-3}
	\jorgen & $\mu + \lambda$ Evolution Strategy~\citep{eiben2015introduction} & CMA-ES~\citep{augerTutorialCMAESEvolution2012} \\ \cline{1-3}
\end{tabular}
	\caption{Optimization algorithms for the design and the control of the \nenvironments co-optimization environments considered in this paper.}
	\label{tab:learning_algorithms}
\end{table}

\subsubsection*{Parameters and experimental setting}

In all of the co-optimization environments, we consider a stopping criterion of $8 \cdot 10^6$ steps.
Each execution is repeated 60 times, and we report the mean and the 95\% confidence band of the mean computed with bootstrap~\citep{arzaComparingTwoSamples2022}.
Besides that, the default parameters of the environments are used with no changes.

\subsection{Experiment 1: Single or two-phase approaches}
\label{sec:exp_single_phase_vs_two_phase}

\begin{figure}[t]
	\centering
	\begin{subfigure}[b]{0.4\textwidth}
		\includegraphics[width=\textwidth]{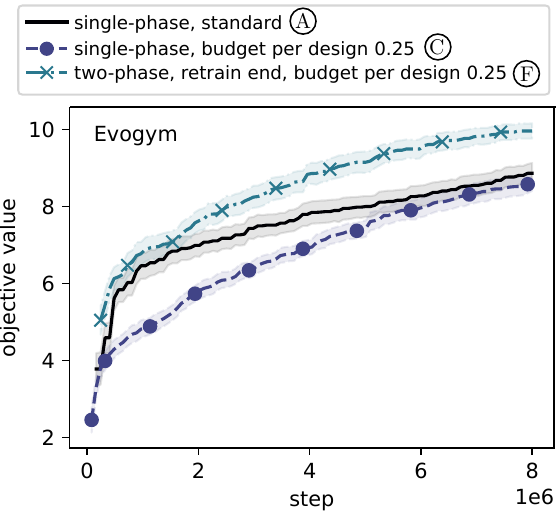}
	\end{subfigure}
	\hfil
	\begin{subfigure}[b]{0.4\textwidth}
		\includegraphics[width=\textwidth]{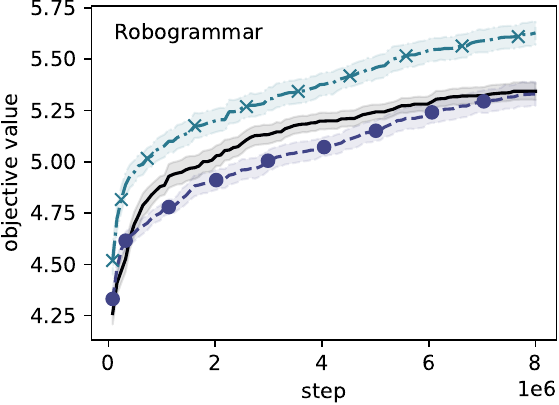}
	\end{subfigure}%
	\vspace{1em}
	\begin{subfigure}[b]{0.4\textwidth}
		\includegraphics[width=\textwidth]{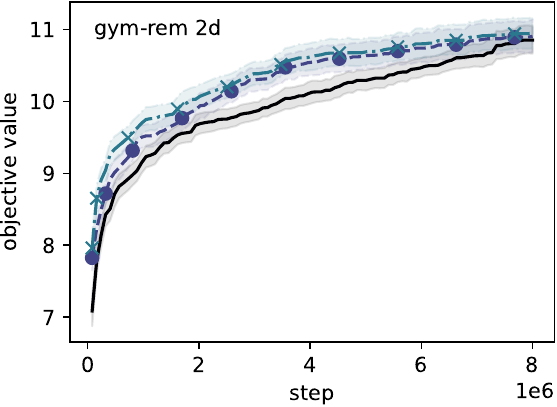}
	\end{subfigure}
	\hfil
	\begin{subfigure}[b]{0.4\textwidth}
		\includegraphics[width=\textwidth]{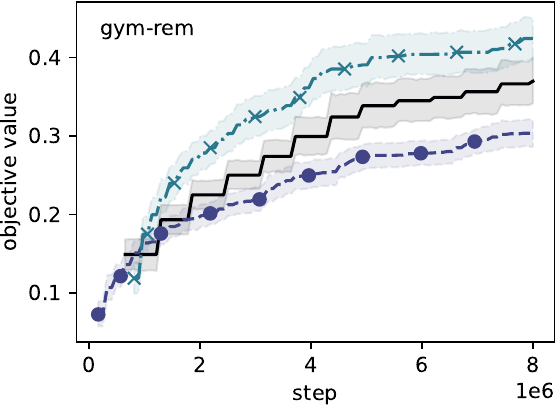}
	\end{subfigure}

	\caption{
        Comparison between \textit{single-phase} and a \textit{two-phase, retrain end} co-optimization of design and control approaches (see 	\textcircled{\scriptsize A}, \textcircled{\scriptsize C} and \textcircled{\scriptsize F} in Figure~\ref{fig:computationalbudgetbarcharts}). 
        The figure shows the mean (line) and the 95\% confidence interval of the mean (shaded area around the line) of 60 repeated executions.
        The confidence interval is computed with bootstrap.
    }
	\label{fig:one_phase_vs_2_phase}
\end{figure}

In this experiment, we compare a \textit{two-phase, retrain end} approach with the standard \textit{single-phase} co-optimization algorithm.
Specifically, we compare the \textit{retrain end} two-phase approach with two different single-phase methods as baselines.
First, we execute the \textit{single-phase, standard} (\textcircled{\scriptsize A} in Figure~\ref{fig:computationalbudgetbarcharts}).
Then, we execute the same \textit{single-phase} procedure, but with a \textit{budget per design} of 0.25 (\textcircled{\scriptsize C} in Figure~\ref{fig:computationalbudgetbarcharts}).
Finally, we compare a \textit{two-phase} approach in which we first execute the \textit{single-phase} algorithm with a \textit{budget per design} of 0.25, followed by a second phase in which only the controller is retrained (\textcircled{\scriptsize F}~in~Figure~\ref{fig:computationalbudgetbarcharts}).
The results of these three executions are shown in Figure~\ref{fig:one_phase_vs_2_phase} for the \nenvironments co-optimization environments considered.

First, observe that in \evogym, \robogrammar, and \jorgen, \textit{single-phase, standard} outperforms \textit{single-phase, budget per design 0.25}.
This means that reducing the training budget per design by $75\%$ is very detrimental to performance.
The \textit{two-phase} approach, in contrast, outperforms the other two methods in all four environments (although the difference is not statistically significant in \veenstra).

\textbf{Discussion}:
The \textit{single-phase, standard} approach evaluates four times fewer designs than \textit{single-phase, budget per design 0.25}, because it uses four times as much resources to train a single design.
However, it still performs worse than \textit{single-phase, standard}.
We theorize this is because, even though more designs can be evaluated, the controllers of the designs are not properly trained.
In other words, even though more designs can be tested, the full potential of the designs cannot be exploited given that the budget allocated to training each design is reduced to 0.25 of the original value.

The \textit{two-phase} approach overcomes the limitations of both single-phase methods: it saves time using less computation budget per design (like the \textit{single-phase, budget per design 0.25}), but avoids its main drawback with a properly trained best design like \textit{single-phase, standard}.

\FloatBarrier
\subsection{Experiment 2: When to retrain the controller}
\label{sec:retrain_end_or_every_new_best}

Two-phased approaches in the literature \citep{rosendoTradeoffMorphologyControl2017,lipsonDifficultyCooptimizingMorphology2016,nygaardOvercomingInitialConvergence2017} consider an initial co-optimization phase in which the design and control are jointly optimized, followed by a second phase in which the design is ``locked'' and only the controller is optimized.
We use \textit{retrain end} in this paper to refer to these methods (see Section~\ref{sec:general_definitions} and \textcircled{\scriptsize E}, \textcircled{\scriptsize F}, \textcircled{\scriptsize G} in Figure~\ref{fig:computationalbudgetbarcharts}).

\begin{figure}[th]
	\centering
	
	\begin{subfigure}[b]{0.40\textwidth}
		\includegraphics[width=\textwidth]{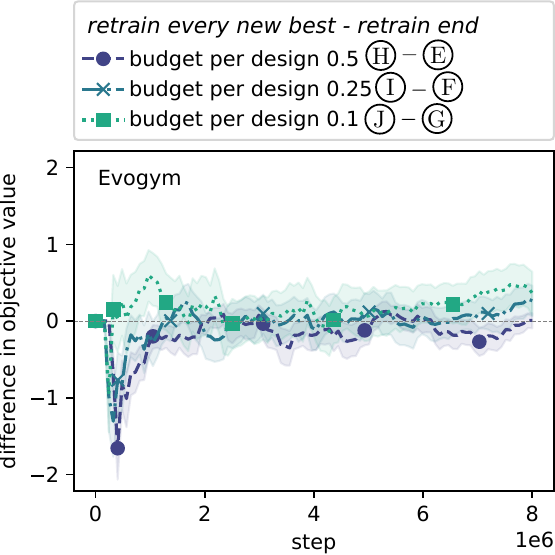}
	\end{subfigure}
	\hfil
	\begin{subfigure}[b]{0.40\textwidth}
		\includegraphics[width=\textwidth]{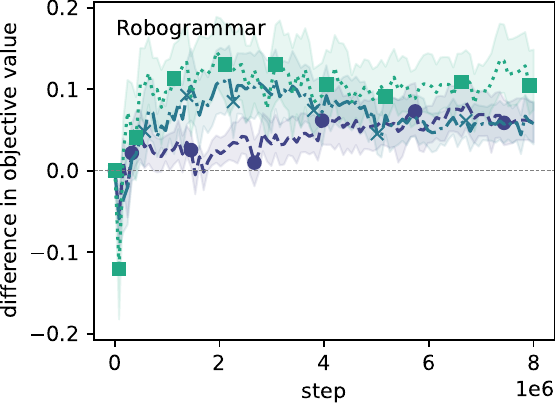}
	\end{subfigure} 

	\vspace{1em}

	\begin{subfigure}[b]{0.40\textwidth}
		\includegraphics[width=\textwidth]{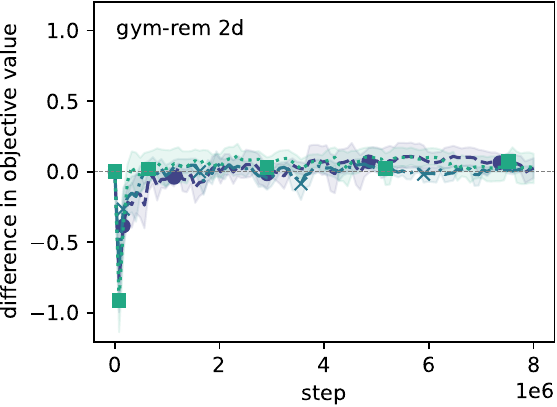}
	\end{subfigure}
	\hfil
	\begin{subfigure}[b]{0.40\textwidth}
		\includegraphics[width=\textwidth]{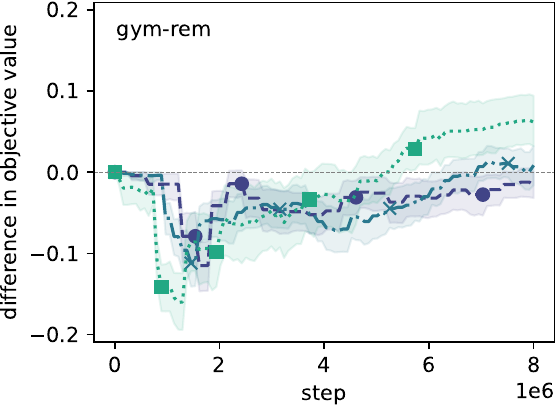}
	\end{subfigure}	
	\caption{Comparison of two-phase co-optimization strategies: \textit{retrain every new best} vs. \textit{retrain end}. A positive difference in objective value means that \textit{retrain every new best} obtains better performance than \textit{retrain end} for the same computation budget. A negative value has the opposite meaning.}
	\label{fig:coopt_strat_comparison}
\end{figure}

However, it is also possible to execute the second ``controller only optimization'' phase each time a new best design is found (instead of only once at the end).
We call this method \textit{retrain every new best}.
In this experiment, we compare \textit{retrain end} and \textit{retrain every new best}, with the same \textit{budget per design} parameter values.
Specifically, we compute the average difference in objective value between these two methods for the same computation budget.
The interpretation is easy: when the difference in objective value is positive, it means that \textit{retrain every new best} performs better than \textit{retrain end}  and vice versa.
The results are shown in Figure~\ref{fig:coopt_strat_comparison}.

Under our experimental setting the results are inconclusive and neither strategy is better than the other for every computation budget and environment.
However, there are some patterns that are common in all environments.
The best strategy for very low computation budgets is \textit{retrain end} for all of the environments, represented by the negative spike for budgets lower than 1e6 steps\footnote{
	This negative spike appears to be delayed in \jorgen. This is to be expected, as \jorgen uses more than 17 times as many steps per design as any of the other environments.
	}.
This result has an intuitive explanation.
At the beginning of the co-design procedure, new best designs are found very frequently, as most of these early designs are generated randomly and do not perform very well.
Consequently, computation budget is wasted if the controllers of these bad designs are retrained, since it is very likely that a better design will be found later on that makes the previous retraining procedure useless.

In addition, we also see that the more we reduce the budget per design, the better the \textit{retrain every new best} seems to perform.
In other words, a lower budget per design gets the most benefit from retraining every new best design instead of only the final best design.
Our hypothesis is that a high reduction in computation budget per design favors \textit{retrain every new best}, because it improves the probability of choosing the optimal design during the co-optimization process significantly more than when the budget is reduced slightly.
We conducted another experiment to verify this hypothesis.

\begin{figure}[h]
	\centering
	\begin{subfigure}[b]{0.40\textwidth}
		\includegraphics[width=\textwidth]{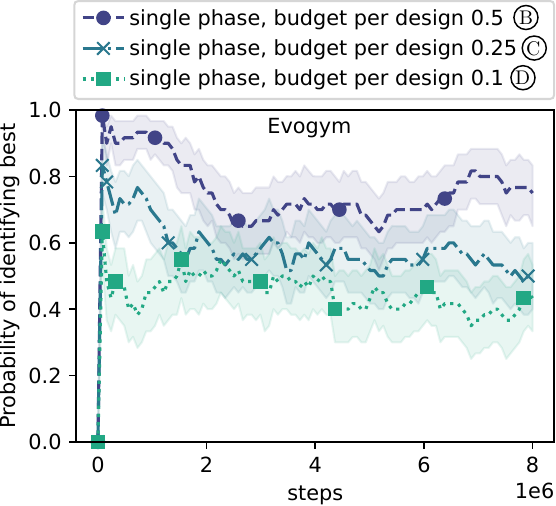}
	\end{subfigure}
	\hfil
	\begin{subfigure}[b]{0.40\textwidth}
		\includegraphics[width=\textwidth]{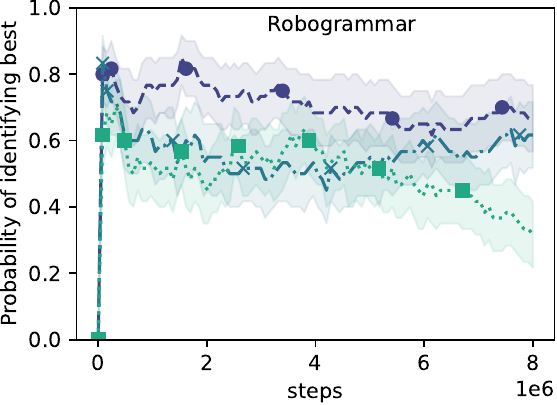}
	\end{subfigure}%

	\vspace{1em}
	
	\begin{subfigure}[b]{0.40\textwidth}
		\includegraphics[width=\textwidth]{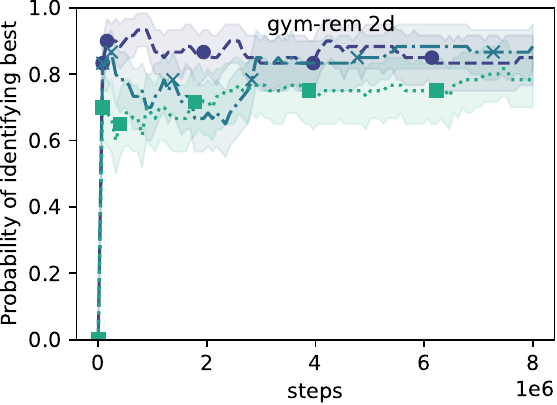}
	\end{subfigure}
	\hfil
	\begin{subfigure}[b]{0.40\textwidth}
		\includegraphics[width=\textwidth]{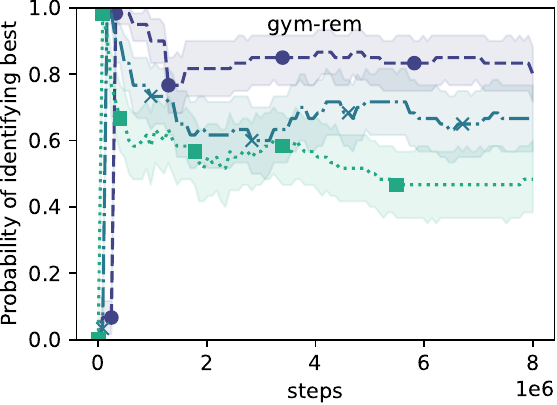}
	\end{subfigure}

	\caption{Probability that the best design found with a reduced computation budget remains the best after retraining with a full budget.}
	\label{fig:prob_retrain_beats_previous}
	\end{figure}

We execute a single-phase co-optimization procedure with reduced budget (\textcircled{\scriptsize B}, \textcircled{\scriptsize C} and \textcircled{\scriptsize D} in Figure~\ref{fig:computationalbudgetbarcharts}), and save each design that was found to be the best so far.
Then, we retrain each of the designs without reducing the computation budget.
Finally, we measured how often the best design is chosen correctly with a reduced budget.
In other words, we estimate the probability that the design chosen as best with the reduced budget still outperforms the rest of the designs when all designs are trained with an additional computation budget.
We show the results in Figure~\ref{fig:prob_retrain_beats_previous}.
As expected, the lowest probability is observed in the figures when the computation budget per design is reduced the most.
We believe that this explains why, in general, a higher benefit from \textit{retrain every new best} is observed when the computation budget per design is lower.

\textbf{Discussion}:
Although \textit{retrain end} is the two-phase approach considered by other works in the literature, the experimentation in this section suggests that \textit{retrain every new best} is also a good performing approach.
Our experiments do not show any consistent difference between the two approaches across environments.
However, \textit{retrain end} showed slightly better performance for very low computation budgets, while \textit{retrain every new best} performed better for a lower computation budget per design (\textit{budget per design} $\leq$ 0.25).

\subsection{Experiment 3: Complexity (number of controllable joints)}
\label{exp:design_size}

\begin{figure}[h]
    \centering
    \begin{subfigure}[b]{0.3\textwidth}
        \includegraphics[width=\textwidth]{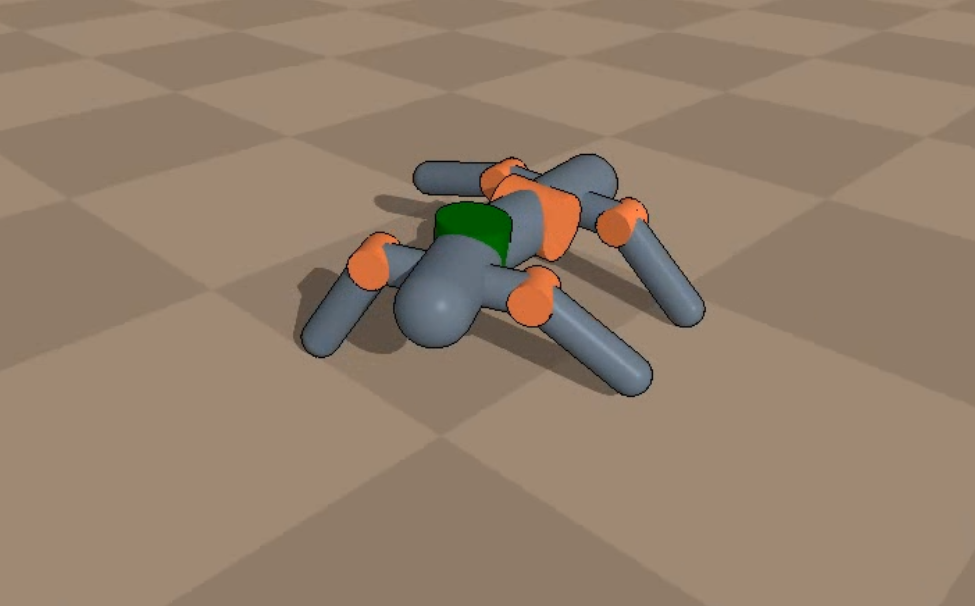}
        \caption{Low Complexity}
        \label{fig:low_complexity}
    \end{subfigure}
    \hfill
    \begin{subfigure}[b]{0.3\textwidth}
        \includegraphics[width=\textwidth]{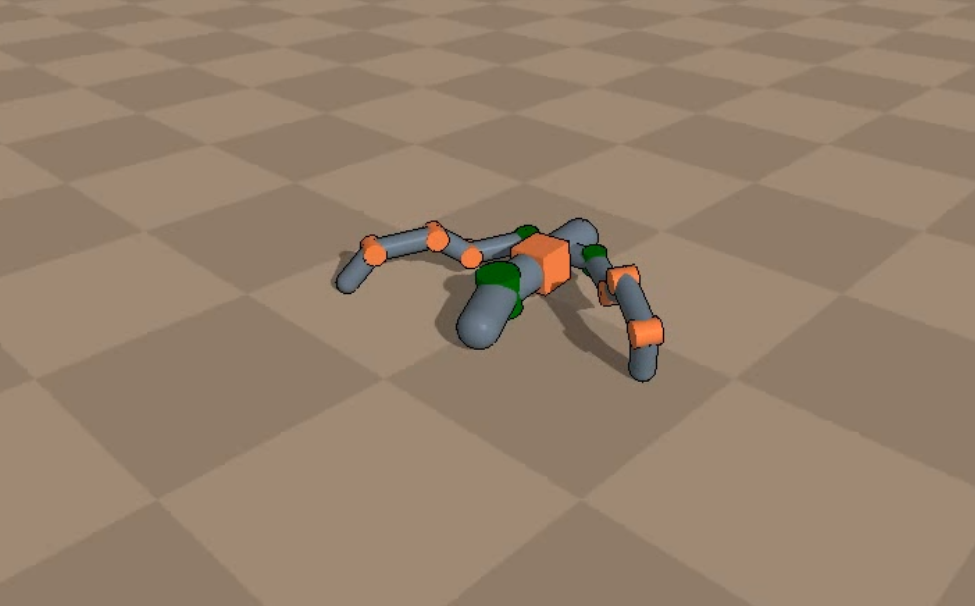}
        \caption{Medium Complexity}
        \label{fig:medium_complexity}
    \end{subfigure}
    \hfill
    \begin{subfigure}[b]{0.3\textwidth}
        \includegraphics[width=\textwidth]{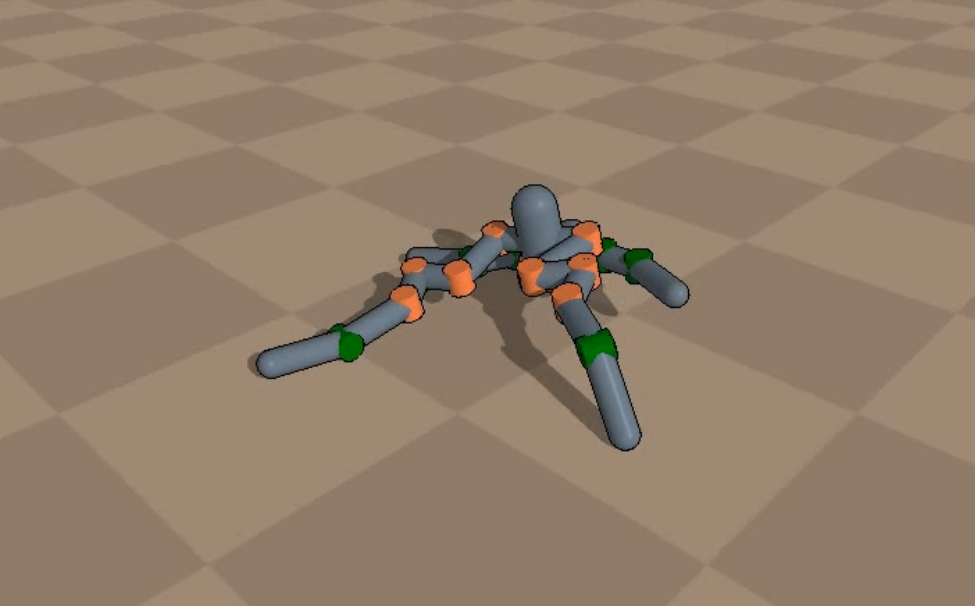}
        \caption{High Complexity}
        \label{fig:high_complexity}
    \end{subfigure}
    \caption{Different levels of complexity. To see how these robots move, we refer the interested reader to \href{https://github.com/EtorArza/NestedOpt/tree/main/animations_for_the_paper/complexity_example}{our GitHub repository}.}
    \label{fig:complexity_levels}
\end{figure}

There is no unique way to measure design complexity \citep{auerbachEnvironmentalInfluenceEvolution2014}.
In this paper, we measure design complexity as the dimensionality of the input space of the control of a design.
In other words, the design complexity is measured as the number of controllable moving parts in the robot (where each moving part needs to be controlled).
In Figure~\ref{fig:complexity_levels}, we show three robots with increasing complexity as an example.
The low complexity robot has 6 control parameters (6 joints); the medium and high complexity robots have 10 and 16 control parameters, respectively.

\begin{figure}[h]
	\centering
	
	\begin{subfigure}[b]{0.40\textwidth}
		\includegraphics[width=\textwidth]{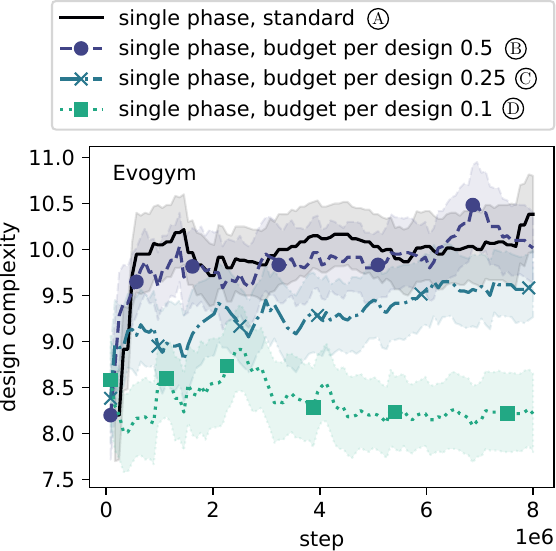}
	\end{subfigure}
	\hfil	
	\begin{subfigure}[b]{0.40\textwidth}
		\includegraphics[width=\textwidth]{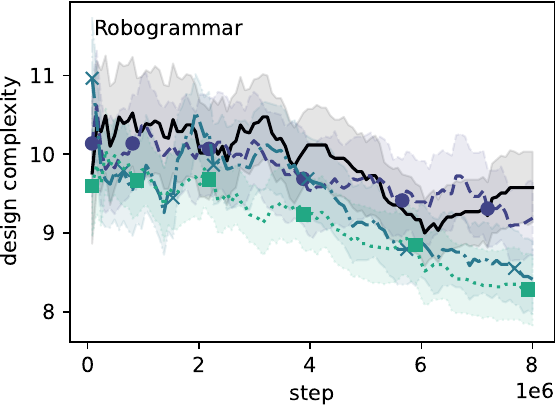}
	\end{subfigure}%
	
	\vspace{1em}
	
	\begin{subfigure}[b]{0.40\textwidth}
		\includegraphics[width=\textwidth]{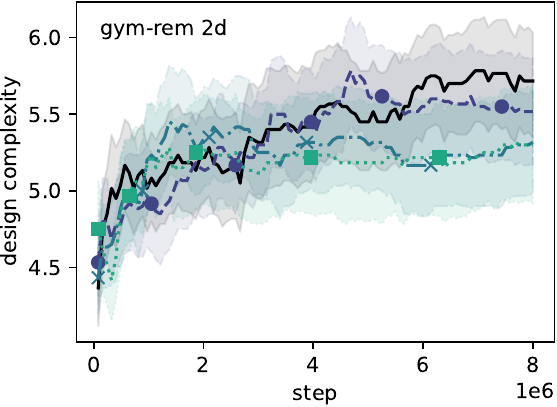}
	\end{subfigure}
	\hfil
	\begin{subfigure}[b]{0.40\textwidth}
		\includegraphics[width=\textwidth]{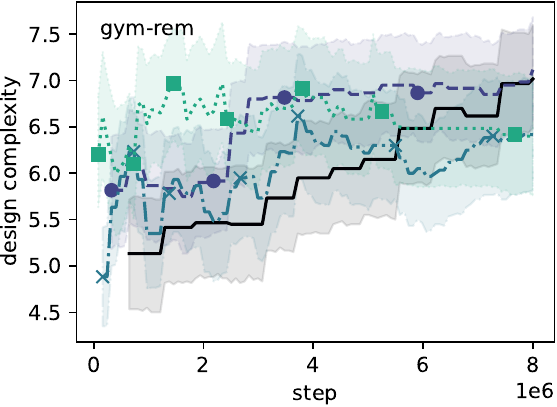}
	\end{subfigure} %

	\caption{The complexity of the best-found design for a given computation budget.}
	\label{fig:reevaleachvsend_morphsize}
\end{figure}

All of the environments considered in the experimentation of this paper do not have a fixed complexity; instead, the complexity of the robots changes during the co-optimization process.
In this experiment, we look into how the training budget influences design complexity.
We study how design complexity changes over time and how it is affected by the computation budget per design.
For each of the four co-optimization environments considered in this paper, in Figure~\ref{fig:reevaleachvsend_morphsize}, we report the design complexity of the best robot found during the co-optimization process and how it changes with respect to reducing the computation budget per design.

In all of the environments except \robogrammar, the design complexity increases with the number of steps.
However, in \robogrammar, the design complexity decreases with the number of steps.
In addition, the lower the number of episodes evaluated per design (lower budget per design parameter), the lower the complexity of the best design seems to be, especially in \robogrammar and \evogym.
In all of the environments, the lowest complexity is observed for \textcircled{\scriptsize D} \textit{budget per design 0.1}, for $8 \cdot10^6$ training steps.

\textbf{Discussion}:
This experiment tried to verify \citet{10.1007/978-3-030-16692-2_41}'s claim that simpler designs are preferred for lower computation budgets, even when the design complexity is allowed to change during the co-optimization procedure.
Our hypothesis for this is that more complex designs require additional training time for the controller to be able to fully exploit the design; hence when less training time is used per design, simpler designs tend to perform better. 
This implies that adjusting the training budget of each design can be used to alter the design complexity of the end result, where a lower budget per design leads to simpler designs.
However, the results are only significant for two out of the four environments, and therefore the results are not conclusive.

\section{Conclusions}
\label{section:conclusions}

In this paper, we experimented on the \textit{from scratch} co-optimization of design and control of robots in environments from the literature.
We studied some of the most discussed topics in the literature of co-optimization of design and control, such as two-phase approaches and design complexity.
In contrast to other works that discuss these topics, our work experiments on four different co-optimization environments from the literature, making the applicability of the observed results more relevant to the general case.

Previous work disagrees on what performs better: a single-phase~\citep{rosendoTradeoffMorphologyControl2017,lipsonDifficultyCooptimizingMorphology2016}, or a two-phase approach~\citep{nygaardOvercomingInitialConvergence2017}.
The first experiment of this paper shows that a two-phase approach---with an initial co-optimization of design and control phase (with a reduced number of episodes per design), followed by a retraining phase of the controller only---outperforms the single-phase approach for the same computation budget.

%

Two-phase co-optimization procedures in the literature~\citep{rosendoTradeoffMorphologyControl2017,lipsonDifficultyCooptimizingMorphology2016,nygaardOvercomingInitialConvergence2017} consider the second ``controller optimization only'' phase after the first co-optimization phase ends.
However, it is also possible to retrain the controller every time a new best design is found.
The second experiment in the paper compares these two variants and shows that retraining the controller only at the end might be preferable for low computation budgets.
Similarly, retraining the control for every new best design might be more effective when the computation budget is large, and also when the controller for each design is trained with a limited number of episodes.

The last experiment of the paper shows that the resources allocated to training each design influence the design complexity.
Design complexity changes as the computation budget increases.
Moreover, the design complexity is also affected by the number of episodes evaluated per design, regardless of the total computation budget.
This means that a straightforward method to obtain simpler designs with a co-optimization algorithm is to reduce the number of episodes evaluated per design (to train each design with less resources during the co-optimization).

Through these experiments, we have investigated challenges inherent in the co-optimization of design and control, such as how to more efficiently distribute the computation budget, design complexity, and two-phased approaches.
The results presented in this paper answer two key questions in the literature.
Firstly, we showed that a two-phase approach with an initial co-optimization phase followed by a controller-only optimization phase outperforms a single-phase co-optimization approach.
Secondly, we analyzed the design complexity and showed how it changes based on the budget allocated to training each of the controllers in the designs, where low computation budget per controller was associated with simpler designs (designs with less controllable joints).
The findings from this paper could be useful for practitioners engaged in the co-optimization of design and control of robots in simulation.
By understanding the relationships between computation budget, design complexity, and two-phased approaches, practitioners can make more informed decisions when proposing co-optimization of design and control problems.

\textbf{Acknowledgments:}


This research is supported by:
\begin{itemize}
\item The Basque Government through the Elkartek program through KONFLOT KK2022/00100
\item Spanish Ministry of Science, Innovation and Universities through PID2019-106453GA-I00/AEI/10.13039/ 501100011033
\item European Commission through the European Union’s Horizon Europe Research and Innovation Programme, under the Grant Agreement No. 101119774 SPEAR
\item The Research Council of Norway through its Centres of Excellence scheme, project number 262762.
\end{itemize}

	
	\bibliographystyle{apalike}
	\bibliography{main}

	\appendix

\end{document}